\documentclass{article} %
\usepackage{cadwam_preprint,times}
\usepackage{graphicx}
\usepackage{booktabs}
\usepackage{tabularx}
\usepackage{array}
\usepackage{multirow}
\usepackage[most]{tcolorbox}

\usepackage{amsmath,amsfonts,bm}

\def\eqref#1{equation~\ref{#1}}

\def\1{\bm{1}}

\DeclareMathAlphabet{\mathsfit}{\encodingdefault}{\sfdefault}{m}{sl}
\SetMathAlphabet{\mathsfit}{bold}{\encodingdefault}{\sfdefault}{bx}{n}

\usepackage{hyperref}
\usepackage{url}
\usepackage{enumitem}

\title{CAD-WAM: Anticipating Geometry for Text-to-CAD Generation}

\author{%
{\small\bfseries
Zhengyang Li$^{1}$ \enspace
Weijie Zheng$^{1}$ \enspace
Yuxin Cui$^{1}$ \enspace
Sijie Wang$^{2}$ \enspace
Wenjie Feng$^{1}$ \enspace
Qiyu Kang$^{1}$\thanks{Correspondence to: Qiyu Kang
\texttt{<qiyukang@ustc.edu.cn>}.}
}\\[0.5em]
{\normalfont\small $^{1}$\,University of Science and Technology of China} \\
{\normalfont\small $^{2}$\,Nanyang Technological University}
}

\hypersetup{
  pdfauthor={Zhengyang Li, Weijie Zheng, Yuxin Cui, Sijie Wang, Wenjie Feng, Qiyu Kang}
}

\hypersetup{pdftitle={CAD-WAM: Anticipating Geometry for Text-to-CAD Generation}}
\usepackage{fancyhdr}

\begin{document}

\setcounter{footnote}{1}
\maketitle

\begin{abstract}
Generating editable computer-aided design (CAD) models from text requires selecting operations that produce the intended changes to the geometry constructed so far. An incorrect operation can compromise subsequent steps, making reliable multi-step generation challenging. Existing methods based on large language models (LLMs) primarily learn to predict commands, without explicit supervision of the geometric states those commands should produce. Inspired by world-action models, we introduce CAD-WAM, which couples geometry-conditioned action generation with next-state prediction. We define complete modeling operations as actions in a CAD domain-specific language (DSL) and encode surface point clouds of intermediate solids into compact latent states using a pretrained variational autoencoder. An LLM backbone generates the next action from the target description and current geometric state, while an auxiliary head on the same backbone predicts the next state to provide geometric supervision during training. We construct step-wise training trajectories from parametric CAD models in the DeepCAD dataset, pairing each trajectory with an expert-level textual description. At inference, each generated action is executed by a CAD kernel, and the resulting solid is re-encoded to inform the next decision. On the DeepCAD test set, CAD-WAM outperforms the task-specific baselines in geometric accuracy and output validity, with larger gains on models requiring more operations, and produces more accurate geometry than general-purpose LLMs prompted with the same DSL. Ablations further show that next-state supervision reduces mean Chamfer distance by 13.2\% relative to an otherwise identical model trained only to predict actions. Code is available at \url{https://github.com/Lzy-1037/CAD-WAM}.
\end{abstract}

\section{Introduction}
Parametric computer-aided design (CAD) represents a model through a sequence of editable geometric operations rather than only its final shape. Generating such models from natural language can reduce the manual effort required for CAD modeling. Existing approaches generate parametric command sequences~\citep{khan2024text2cad} or fine-tune large language models (LLMs) to produce executable programs in CAD scripting languages such as CadQuery and FeatureScript~\citep{guan2025cadcoder,pyatov2026cadfs,kolodiazhnyi2026cadrille}.

Reliable multi-step CAD generation requires more than producing executable commands: each operation must also realize the intended change to the geometry constructed so far. An incorrect operation can alter the solid on which subsequent operations act, causing further deviations from the target. Our experiments also reveal an uneven error distribution: several baselines achieve low median geometric error, yet some outputs deviate substantially from the target (Figure~\ref{fig:results}). Existing methods incorporate geometric information through reinforcement learning with geometric rewards~\citep{guan2025cadcoder,kolodiazhnyi2026cadrille} or step-wise generation conditioned on the current boundary representation (B-rep)~\citep{qi2026pointercad}. These approaches use geometry to evaluate generated programs or inform the next decision, but do not explicitly supervise predictions of the next geometric state. This motivates learning each modeling step through both its symbolic action and its intended geometric outcome, connecting operation generation with the evolving shape of the CAD model.

World-action models provide a natural perspective on this formulation by coupling action learning with prediction of future observations~\citep{tian2025seer,zhu2025uwm}. Future prediction can provide additional supervision during training without making action generation depend on predicted states at inference~\citep{ye2026gigaworld}. CAD offers a direct source of such supervision: executing expert programs yields the intermediate solids after each operation. These geometric outcomes can supervise the representation used for action generation, while the CAD kernel provides the actual geometry for subsequent decisions during inference.

Building on this idea, we introduce CAD-WAM, which jointly learns next-action generation and next-state prediction from the target description and current geometry. We design a CAD domain-specific language (DSL) that groups a sketch, its extrusion, and the associated Boolean operation into a complete modeling action, yielding a solid after each successful execution. We pretrain SetVAE~\citep{kim2021setvae}, a variational autoencoder for point clouds, on surface point clouds of intermediate and final solids from the training split. Its encoder is then frozen to provide compact latent states that serve both as geometric inputs and as targets for next-state prediction. The Geometric State Embedding maps the current state to geometric tokens, which are processed together with textual tokens by a shared LLM backbone. The Action Head generates the next DSL action, while the State Prediction Head regresses the next latent state. Joint training allows next-state supervision to shape the shared representation used for action generation. At inference, each generated action is executed by a CAD kernel, and the resulting solid is re-encoded to condition the next decision.

To train CAD-WAM, we construct step-wise trajectories from DeepCAD, a dataset of parametric CAD models~\citep{wu2021deepcad}, by converting their construction sequences into our DSL and executing them operation by operation. Each trajectory is paired with an expert-level textual description released with Pointer-CAD~\citep{qi2026pointercad}. At each step, the description and current geometric state form the input, while the next action and resulting geometric state provide supervision.
We compare CAD-WAM with task-specific baselines and general-purpose LLMs on the DeepCAD test set, applying a common geometric evaluation protocol to all available outputs. CAD-WAM achieves higher geometric accuracy and lower output invalidity than the evaluated task-specific baselines, with larger gains on models requiring more operations. It also produces more accurate geometry than general-purpose LLMs prompted with the same DSL. Ablations show that the latent-state model outperforms its textual-state counterpart and that next-state supervision reduces mean Chamfer distance by 13.2\% relative to an otherwise identical model trained only to predict actions.

Our contributions are summarized as follows:
\begin{itemize}[label=$\bullet$, topsep=0pt, itemsep=3pt, partopsep=0pt, parsep=0pt,leftmargin=10pt]
    \item We introduce CAD-WAM, a text-to-CAD framework that couples geometry-conditioned action generation with next-state prediction, while using kernel-executed geometry as feedback during inference.
    \item We develop an operation-level CAD DSL and a compact latent state representation, and construct step-wise training data that pair executable modeling actions with their geometric outcomes.
    \item We demonstrate improved geometric accuracy over the evaluated task-specific and general-purpose models on the DeepCAD dataset, and establish the benefit of next-state supervision through controlled ablations.
\end{itemize}

\section{Related Work}
\subsection{CAD Generation with Language Models}
  \label{sec:rw_cad}

  Learning-based CAD generation is usually formulated over the construction sequence rather than the final
  geometry.
  DeepCAD introduced this formulation together with a dataset of 178{,}238 sketch-and-extrude
  sequences~\citep{wu2021deepcad}, and later models improved generation and controllability over the same
  representation without language supervision~\citep{xu2022skexgen,xu2023hnccad}.
  Language models then replaced the task-specific command vocabulary by executable code: CAD-Recode
  reverse engineers CadQuery programs from point clouds~\citep{rukhovich2025cadrecode}, and cadrille
  extends this setting to point clouds, images and text with reinforcement
  learning~\citep{kolodiazhnyi2026cadrille}.
  For text-conditioned generation, models emit either command
sequences~\citep{khan2024text2cad,govindarajan2025cadmium} or programs in
  CadQuery~\citep{guan2025cadcoder} and FeatureScript~\citep{pyatov2026cadfs}.
  Closest to our setting, Pointer-CAD generates a model step by step and conditions each step on the
  boundary representation obtained so far~\citep{qi2026pointercad}.
  Unlike these methods, which either ignore the intermediate geometry or merely condition on it, our work
  encodes it as a latent state and predicts the state that the next action produces.
\subsection{World Models and World-Action Models}
  \label{sec:rw_wm}

  World models learn the dynamics of an environment in a latent space, so that the effect of an action can
  be estimated without executing it~\citep{ha2018world,hafner2020dream}.
  The idea has since been scaled to generative video models, which synthesize future observations
  conditioned on text or on latent actions~\citep{du2023unipi,bruce2024genie}.

  Predicting future observations is useful for control only when it informs the choice of action.
  World-action models therefore couple prediction with action generation, and differ in how the two are
  combined: the action may be inferred from a predicted future observation~\citep{tian2025seer}, actions
  and observations may be produced within a single generative process~\citep{zhu2025uwm}, or future
  prediction may serve as an auxiliary objective during training~\citep{ye2026gigaworld}.
  CAD-WAM draws on this idea and applies it to CAD modeling, where the transition induced by an
    action is deterministic and the resulting geometry can be obtained exactly with a CAD kernel.
  The predicted state is therefore a well-defined target, and the observed state requires no
  approximation.

\section{CAD-WAM}
\subsection{Problem Formulation}
\label{sec:formulation}

Given a natural-language description $g$ of a target CAD model,
CAD-WAM constructs the model in a step-by-step manner through
a sequence of state-dependent decisions. We design a CAD domain-specific language (DSL;
Section~\ref{sec:dsl}) that represents CAD operations
(e.g., extrusions and Boolean operations) as executable actions.
The construction process is thus expressed as an action sequence
$(a_0,a_1,\dots,a_{T-1})$, where $a_t \in \mathcal{A}$ and
$\mathcal{A}$ denotes the action space defined by the DSL.
Let $S_t$ denote the B-rep of the
intermediate CAD model after $t$ actions.
Each action updates the current geometry through a
deterministic transition implemented by the CAD kernel:
\begin{equation}
S_{t+1} = \mathcal{K}(S_t, a_t),
\qquad t = 0,\ldots,T-1,
\label{eq:transition}
\end{equation}
where $\mathcal{K}$ denotes the CAD kernel's execution function.

We represent each nonempty intermediate model by a compact
geometric state. Specifically, we sample a point cloud
$\mathcal{P}(S_t)$ from its surface and encode it using a
SetVAE encoder $E_\phi$~\citep{kim2021setvae}
(Section~\ref{sec:state}):
\begin{equation}
\mathbf{z}_t =
E_\phi\big(\mathcal{P}(S_t)\big)
\in \mathbb{R}^{1024}.
\label{eq:state}
\end{equation}
The process starts from an empty model $S_0$, represented
by the zero vector $\mathbf{z}_0=\mathbf{0}$.

CAD-WAM conditions each decision on the textual target and
the current geometric state, without explicitly taking the
action history as input. Given $(g,\mathbf{z}_t)$, it jointly
predicts the next action and the resulting geometric state
by modeling
\begin{equation}
p_\theta\big(a_t,\mathbf{z}_{t+1}
\mid g,\mathbf{z}_t\big),
\label{eq:joint}
\end{equation}
where $\theta$ denotes the parameters of CAD-WAM.
This formulation couples action generation with prediction
of the resulting geometric state, encouraging the model to
anticipate the geometric consequences of the next action.

\subsection{CAD Domain-Specific Language}
  \label{sec:dsl}
  \begin{figure}[t]
    \centering
    \includegraphics[width=\linewidth]{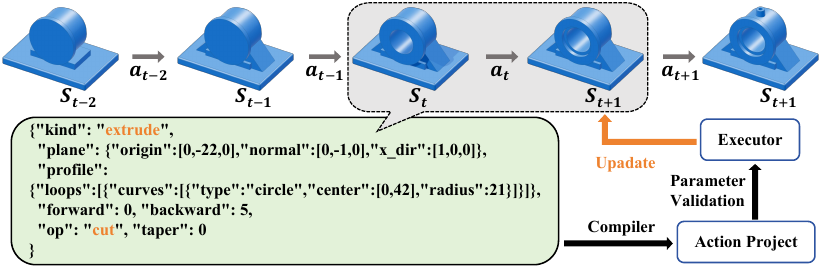}
    \caption{Our CAD DSL. Top: a construction sequence in which every action yields a valid solid. Bottom
  left: an action serialized as a JSON object. Bottom right: execution of an action, whose parameters are
  validated before the CAD kernel is invoked.}
    \label{fig:dsl}
\end{figure}
We design a CAD DSL in which each modeling action produces
a CAD solid upon successful execution, providing an explicit geometric state after each modeling step. Figure~\ref{fig:dsl} shows a construction sequence in this DSL, the JSON form of an action, and the
  procedure by which an action is executed.
In contrast, a single command token~\citep{wu2021deepcad} or code statement~\citep{rukhovich2025cadrecode,guan2025cadcoder} does not
  necessarily yield a solid, which leaves the intermediate geometry difficult to represent.
The DSL comprises a geometric layer, which defines planar profiles on sketch planes as closed loops of
  lines, arcs, and circles with nested loops forming holes, and an action layer built on these profiles.
  Each action is a one-sided, symmetric, or two-sided sketch-based extrusion, with its Boolean operation
  on the current model as a parameter.

  Actions are serialized as JSON objects with continuous parameters, and a dedicated end action
  terminates the construction. We also check the parameters of each action before
  passing it to the CAD kernel~\citep{cadquery}.
  The complete grammar and validation rules are provided in Appendix~\ref{app:dsl}.
\subsection{Geometric State Representation}
  \label{sec:state}

  World models commonly encode high-dimensional observations into a compact latent space with a
  variational autoencoder (VAE) and model dynamics within that space~\citep{ha2018world,hafner2020dream}.
  Inspired by this practice, we represent each intermediate model $S_t$ by a latent geometric state.
  We observe $S_t$ through a point cloud sampled from its surface, which captures the shape of the solid
  in a unified form for models of arbitrary complexity.
  We encode this point cloud with SetVAE~\citep{kim2021setvae}, a permutation-invariant VAE for sets,
  into a compact fixed-dimensional state.
  Its KL-regularized latent space provides a smooth target for state prediction.

  To construct $\mathcal{P}(S_t)$, we sample 2{,}048 points uniformly over the surface of $S_t$.
  We then center them at the bounding-box center and scale them to a fixed extent, so that intermediate
  solids of very different sizes are encoded at a comparable scale.
  Since this normalization removes absolute position and scale, we supply the size and center of the
  bounding box of $S_t$ to the model as part of its textual input.
  SetVAE encodes a set through a hierarchy of seven latent levels, which contain 1, 1, 2, 4, 8, 16,
  and 32 latent vectors of dimension 16.
  We concatenate the posterior means of all levels to obtain the deterministic geometric state
  $\mathbf{z}_t\in\mathbb{R}^{1024}$.
 We pretrain SetVAE on the point clouds of all intermediate and final solids in the training split
  (Section~\ref{sec:data}) and use its encoder as $E_\phi$, so that the encoder covers the partial
  models encountered as states.
  Training details and reconstruction quality are reported in Appendix~\ref{app:setvae}.
\subsection{Model Architecture}
  \label{sec:arch}
  \begin{figure}[t]
    \centering
    \includegraphics[width=\linewidth]{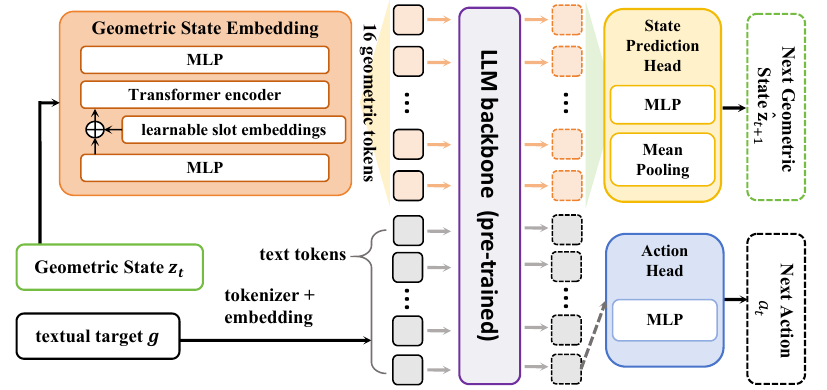}
    \caption{Architecture of CAD-WAM: the Geometric State Embedding maps the geometric state to 16
    geometric tokens that follow the textual tokens in the backbone; the Action Head generates the next
    action and the State Prediction Head regresses the next geometric state.}
    \label{fig:arch}
\end{figure}
  Figure~\ref{fig:arch} illustrates the architecture of CAD-WAM, which realizes the joint distribution in
  Eq.~(\ref{eq:joint}) with a shared LLM backbone and two prediction branches.
  At step $t$, the Geometric State Embedding converts the geometric state $\mathbf{z}_t$ into $K=16$
  geometric tokens.
  These tokens are integrated with the textual tokens and jointly processed by the backbone.
  Its hidden states are then decoded along two branches: the Action Head generates the next action $a_t$,
  and the State Prediction Head regresses the next geometric state $\hat{\mathbf{z}}_{t+1}$.
  In addition to the pretrained LLM, CAD-WAM introduces only the Geometric State Embedding and the State
  Prediction Head, which together contain 11.4M parameters, fewer than 1\% of a 1.5B-parameter backbone.

  \paragraph{Geometric State Embedding.}
  Multimodal LLMs commonly expose a continuous representation to the backbone as a short sequence of
  tokens~\citep{alayrac2022flamingo,li2023blip2}.
  Following this practice, we design the Geometric State Embedding, which maps the geometric state
  $\mathbf{z}_t$ to $K$ geometric tokens in the embedding space of the backbone.
  The module first normalizes $\mathbf{z}_t$ and expands it with $\mathbf{W}_{1}$ into $K$ slots of width
  $d_{g}$, to each of which a learnable slot embedding in $\mathbf{P}$ is added.
  A two-layer bidirectional Transformer encoder $f_{g}$ then allows the slots to exchange information,
  after which the projection $\mathbf{W}_{2}$ maps them to the hidden size $d$ of the backbone:
  \begin{equation}
    \mathbf{G}_t=\operatorname{LN}\Big(f_{g}\big(\operatorname{reshape}(\mathbf{W}_{1}\operatorname{LN}(\mathbf{z}_t))
    +\mathbf{P}\big)\Big)\mathbf{W}_{2}\in\mathbb{R}^{K\times d},
    \label{eq:gse}
  \end{equation}
  where $\mathbf{W}_{1}\in\mathbb{R}^{Kd_{g}\times1024}$, $\mathbf{P}\in\mathbb{R}^{K\times d_{g}}$,
  $\mathbf{W}_{2}\in\mathbb{R}^{d_{g}\times d}$, $\operatorname{LN}$ denotes layer normalization, and
  biases are omitted for brevity.

  \paragraph{Shared Backbone and Action Head.}
  At each step, the token embeddings of the textual input, which contains the textual target $g$, are
  followed by the $K$ geometric tokens $\mathbf{G}_t$, and together they form the input sequence of the
  backbone.
  Placing the geometric tokens after the text allows their hidden states to attend to the entire textual
  input under causal attention.
  The resulting sequence is processed by a pretrained decoder-only LLM, which we leave architecturally
  unchanged.
  The Action Head is the pretrained language-modeling head of this backbone, and it generates the next
  action $a_t$ autoregressively as text in our DSL.
  Since actions are expressed as text, this head can be reused directly, thereby preserving the generative
  capability acquired during pretraining.

  \paragraph{State Prediction Head.}
  We design the State Prediction Head to predict the geometric state that the next action will produce.
  Let $\mathbf{H}_t \in \mathbb{R}^{K \times d}$ denote the final hidden states at the $K$ geometric
  positions, which encode the textual target together with the current geometric state.
  The head averages these hidden states, normalizes the result, and regresses the next geometric state
  through a two-layer MLP with a GELU activation:
  \begin{equation}
  \hat{\mathbf{z}}_{t+1} = \mathrm{MLP}\left( \mathrm{LN}\left( \frac{1}{K} \sum_{k=1}^{K} \mathbf{H}_{t,k} \right) \right) \in \mathbb{R}^{1024}.
  \label{eq:sph}
  \end{equation}
  The State Prediction Head takes $\mathbf{H}_t$ as its input, whereas the action tokens are decoded from
  the subsequent positions, which attend to the geometric tokens under causal attention.
  CAD-WAM thus realizes the joint prediction in Eq.~(\ref{eq:joint}) within a single model, in which both
  branches share the backbone and the Geometric State Embedding.
  Since gradients from both objectives propagate into these shared modules, action generation and state prediction jointly shape a common representation and thereby influence each other. Complete architectural details and hyperparameters are provided in Appendix~\ref{app:impl}.

\subsection{Training and Closed-Loop Inference}
  \label{sec:training}

  \paragraph{Training.}
  We train CAD-WAM on expert trajectories, while the encoder $E_\phi$ obtained in Section~\ref{sec:state}
  remains frozen, so that the regression target of state prediction stays fixed.
  Every step of a trajectory provides one supervised example: the geometric state $\mathbf{z}_t$ is
  obtained by encoding the solid that results from executing the first $t$ ground-truth actions, and the
  supervision consists of the next action $a_t$ and the next geometric state $\mathbf{z}_{t+1}$.
  Since all states are precomputed, the steps of a trajectory form independent examples and are processed
  in parallel.
  The two branches are trained jointly with
  \begin{equation}
  \mathcal{L} = -\sum_{i} \log \pi_\theta \big( a_t^{(i)} \mid a_t^{(<i)}, g, \mathbf{z}_t \big)
  + \frac{\lambda}{D} \big\| \hat{\mathbf{z}}_{t+1} - \mathbf{z}_{t+1} \big\|_2^2 ,
  \label{eq:loss}
  \end{equation}
  where $a_t^{(i)}$ is the $i$-th token of $a_t$ and $D = 1024$ is the dimension of the geometric state.
  The objective combines two terms: a cross-entropy loss over the action tokens, which trains the Action
  Head, and a mean squared error between the predicted and the target geometric state, which trains the
  State Prediction Head.
  The weight $\lambda$ balances the two terms.
  \paragraph{Closed-Loop Inference.}
  \begin{figure}[t]
    \centering
    \includegraphics[width=\linewidth]{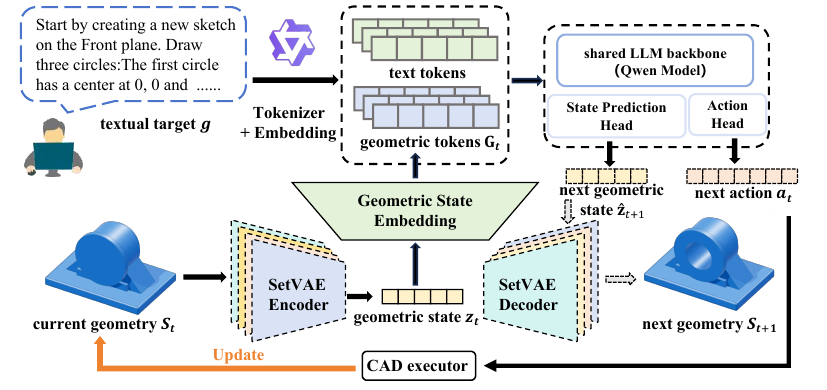}
    \caption{Overview of CAD-WAM, which builds the model step by step, executing each predicted action and
    re-encoding the resulting solid.}
    \label{fig:pipeline}
\end{figure}
  Figure~\ref{fig:pipeline} gives an overview of the complete pipeline.
  At inference, CAD-WAM constructs the CAD model step by step.
  At each step, the model receives the textual target $g$ together with the geometric state
  $\mathbf{z}_t$ of the current model, and predicts the next action $a_t$ and the next geometric state
  $\hat{\mathbf{z}}_{t+1}$.
  The predicted action is then executed by the CAD kernel, and the resulting solid is encoded by $E_\phi$
  into $\mathbf{z}_{t+1}$, which the model receives at the next step.
  The loop is thus closed through the kernel and the encoder, so the state observed by the model always
  reflects the geometry that has actually been built.
  Generation terminates when the model emits the end action, when an action fails verification, or when a
  budget of $T_{\max}$ actions is reached, and the resulting solid is returned.
\section{Step-Wise CAD Dataset}
  \label{sec:data}

  We construct a step-wise dataset from DeepCAD~\citep{wu2021deepcad}, a collection of 178{,}238
  parametric CAD models, paired with the expert-level textual descriptions provided by
  Recap-OmniCAD~\citep{qi2026pointercad}, which serve as the textual targets $g$.
  We adopt the official DeepCAD split throughout.

  Each DeepCAD program is converted into a sequence of actions in our DSL, and this sequence is executed
  step by step to obtain the intermediate models $S_0,\dots,S_T$.
  Every intermediate model is sampled and encoded into its geometric state with the pretrained encoder
  $E_\phi$ (Section~\ref{sec:state}).
  Each step then yields one record: the textual target $g$ and the current geometric state $\mathbf{z}_t$
  form the input, while the next action $a_t$ and the next geometric state $\mathbf{z}_{t+1}$ provide the
  supervision.
  A final record supervising the end action completes each trajectory.

  We retain a model only if it is annotated with a description and its executed geometry matches the
  reference model released with DeepCAD within a geometric tolerance.
  This yields 174{,}582 models and 526{,}010 records, of which 157{,}945 models with 475{,}656 records are
  used for training and 8{,}755 models with 26{,}394 records for validation.
  For evaluation, we use all 7{,}962 test models annotated with a description. Statistics of the resulting programs and an example record are provided in Appendix~\ref{app:data}.
\section{Experiments}
\subsection{Experimental Setup}
  \label{sec:setup}

  \paragraph{Evaluation Set.}
  We evaluate on all 7{,}962 test models of the official DeepCAD split that are annotated with a textual
  description (Section~\ref{sec:data}), and compare each generated model against the reference solid
  released with DeepCAD.

  \paragraph{Baselines.}
  We compare CAD-WAM with two groups of methods.
  The first consists of task-specific text-to-CAD models: Text2CAD~\citep{khan2024text2cad},
  CAD-Coder~\citep{guan2025cadcoder}, CADFS~\citep{pyatov2026cadfs},
  cadrille~\citep{kolodiazhnyi2026cadrille}, and Pointer-CAD~\citep{qi2026pointercad}.
  The second comprises general-purpose LLMs that generate CAD programs without task-specific training,
  namely Qwen3.8-Flash~\citep{qwen2026qwen38flash},
  DeepSeek-V4.1-Flash~\citep{deepseek2026v41flash}, GPT-6 Astra~\citep{openai2026gpt6astra}, and
  Claude Opus 5.5~\citep{anthropic2026opus55}.
  As these general-purpose LLMs are not trained for this task, we provide the DSL specification in the prompt and
  return three orthographic views of the solid built so far after each action.
  We generate the outputs of Text2CAD, CAD-Coder, and cadrille from their published checkpoints.
  For CADFS, we evaluate the inference results of the Qwen3-8B variant reported in its ablation study.
  For Pointer-CAD, we directly report its published results.
  Details of the baseline are provided in Appendix~\ref{app:baselines}.
  \paragraph{Metrics and Protocol.}
  Results reported in previous work are not directly comparable, since their evaluation protocols differ
  in the normalization of the solids (e.g., $[-1,1]^3$ or $[-0.5,0.5]^3$) and in the number of
  sampled points (from 8{,}192 to 100{,}000), among other aspects.
  We therefore re-score all systems under a unified protocol, in which every generated and reference solid
  is meshed, normalized, and sampled identically.
  Each task-specific baseline is evaluated with the textual annotations on which it was trained. All of them generate models of the same benchmark, since their training data derive from DeepCAD.
  The general-purpose LLMs are not trained for this task, and we therefore prompt them with the
  descriptions used in this work to generate DSL code, which is executed to obtain the corresponding
  solid.
  We adopt a comprehensive set of metrics covering both the validity and the geometric fidelity of the
  generated models, which are defined in detail in Appendix~\ref{app:eval}.
  \paragraph{Implementation Details.}
  CAD-WAM is built on Qwen2.5-1.5B-Instruct~\citep{qwen2024qwen25}, whose parameters are fully fine-tuned together
  with the Geometric State Embedding and the State Prediction Head, with $\lambda = 0.03$.
  The encoder $E_\phi$ is pretrained as described in Section~\ref{sec:state} and remains frozen.
  At inference, actions are decoded greedily, and generation is limited to $T_{\max}=20$ actions.
  Training hyperparameters, hardware, and further details are reported in Appendix~\ref{app:impl}.

  \begin{table}[t]
    \centering
    \caption{
    Quantitative comparison with existing CAD generation methods
    and general-purpose large language models.
    }
    \label{tab:quantitative_comparison}

    \small
    \setlength{\tabcolsep}{1.6pt}
    \renewcommand{\arraystretch}{1.15}
    \renewcommand{\tabularxcolumn}[1]{m{#1}}

    \begin{tabularx}{\linewidth}{
    @{}
    >{\centering\arraybackslash}m{0.16\linewidth}
    *{6}{>{\centering\arraybackslash}X}
    |
    *{4}{>{\centering\arraybackslash}X}
    @{}
    }
    \toprule
    \textbf{Metric}
    & {\scriptsize\textbf{\shortstack[c]{Text2\\CAD}}}
    & {\scriptsize\textbf{\shortstack[c]{CAD-\\Coder}}}
    & {\scriptsize\textbf{\shortstack[c]{CADFS-\\8B}}}
    & {\scriptsize\textbf{\shortstack[c]{Pointer-\\CAD}}}
    & {\scriptsize\textbf{\shortstack[c]{cadrille\\(text)}}}
    & {\scriptsize\textbf{\shortstack[c]{CAD-\\WAM\\(Ours)}}}
    & {\scriptsize\textbf{\shortstack[c]{Qwen3.8-\\Flash}}}
    & {\scriptsize\textbf{\shortstack[c]{DeepSeek\\V4.1-\\Flash}}}
    & {\scriptsize\textbf{\shortstack[c]{GPT-6\\Astra}}}
    & {\scriptsize\textbf{\shortstack[c]{Claude\\Opus 5.5}}}
    \\

    \midrule

    IR (\%) $\downarrow$
    & 2.627
    & 0.741
    & 8.125
    & 8.800
    & 2.529
    & 0.666
    & 28.360
    & 29.478
    & \textbf{0.025}
    & 9.068
    \\

    \multirow{2}{*}{
    \begin{tabular}{@{}c@{\hspace{5pt}}l@{}}
    \multirow{2}{*}{CD $\downarrow$}
    & \textit{mean} \\
    & \textit{median}
    \end{tabular}
    }
    & 12.896
    & 3.292
    & 4.151
    & 2.580
    & 3.391
    & \textbf{1.142}
    & 3.717
    & 3.158
    & 1.755
    & 3.106
    \\

    &
    0.554
    & 0.210
    & 0.196
    & 0.300
    & 0.218
    & \textbf{0.166}
    & 0.206
    & 0.200
    & 0.173
    & 0.197
    \\

    \addlinespace[2pt]

    \multirow{2}{*}{
    \begin{tabular}{@{}c@{\hspace{5pt}}l@{}}
    \multirow{2}{*}{HD$\downarrow$}
    & \textit{mean} \\
    & \textit{median}
    \end{tabular}
    }
    & 14.446
    & 6.576
    & 7.491
    & --
    & 7.365
    & \textbf{4.134}
    & 6.634
    & 6.304
    & 4.728
    & 6.354
    \\

    &
    5.687
    & 3.089
    & 3.038
    & --
    & 3.166
    & \textbf{2.783}
    & 3.053
    & 3.029
    & 2.841
    & 3.020
    \\

    \addlinespace[2pt]

    \multirow{2}{*}{
    \begin{tabular}{@{}c@{\hspace{5pt}}l@{}}
    \multirow{2}{*}{NC $\uparrow$}
    & \textit{mean} \\
    & \textit{median}
    \end{tabular}
    }
    & 0.693
    & 0.845
    & 0.839
    & --
    & 0.836
    & \textbf{0.903}
    & 0.854
    & 0.860
    & 0.888
    & 0.859
    \\

    &
    0.840
    & 0.960
    & 0.957
    & --
    & 0.960
    & \textbf{0.964}
    & 0.962
    & 0.963
    & 0.962
    & 0.961
    \\

    \addlinespace[2pt]

    ECD $\downarrow$
    & 45.301
    & 12.289
    & 11.606
    & --
    & 14.128
    & \textbf{7.313}
    & 13.137
    & 12.124
    & 8.186
    & 11.112
    \\

    IoU $\uparrow$
    & 0.631
    & 0.877
    & 0.879
    & --
    & 0.843
    & \textbf{0.964}
    & 0.875
    & 0.889
    & 0.931
    & 0.891
    \\

    \shortstack[c]{Euler match\\(\%) $\uparrow$}
    & 80.712
    & 91.596
    & 85.960
    & --
    & 86.706
    & \textbf{95.850}
    & 92.089
    & 92.840
    & 95.434
    & 91.879
    \\

    \bottomrule
    \end{tabularx}
    \end{table}

\begin{figure}[t]
  \centering
  \includegraphics[width=\linewidth]{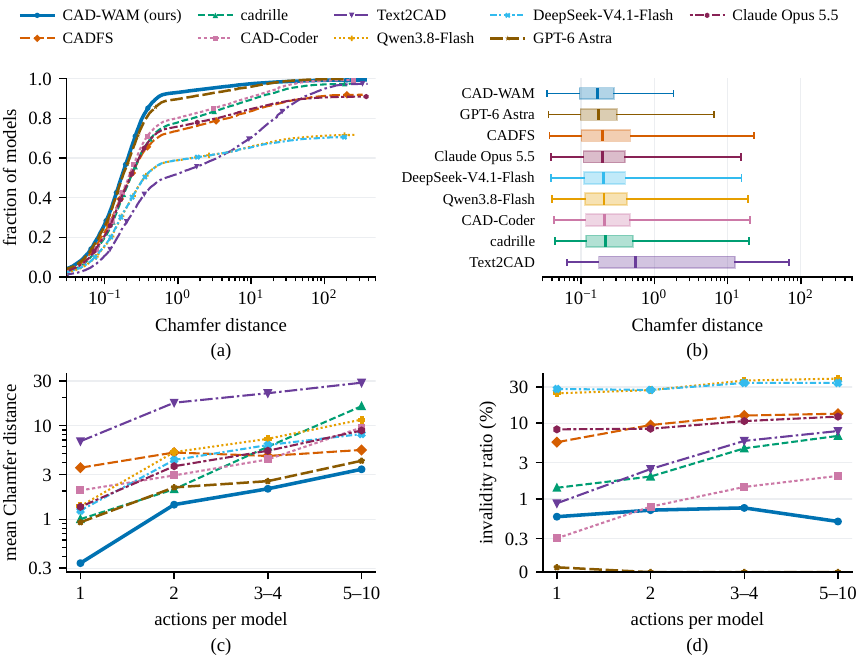}
  \caption{Per-model results for all compared systems. (a) Cumulative distribution of the Chamfer
    distance over the attempted models. (b) Quartile boxes with p5--p95 whiskers, ordered by median.
    (c) Mean Chamfer distance by number of actions. (d) Invalidity ratio by number of actions.
    Pointer-CAD is omitted, as its predictions are not accessible to us.}
  \label{fig:results}
  \end{figure}
 \subsection{Comparison with Existing Methods}
  \label{sec:main_results}

  Table~\ref{tab:quantitative_comparison} compares CAD-WAM with existing CAD generation methods and with
  general-purpose LLMs under the unified evaluation protocol.
  CAD-WAM attains the best value on every metric except the invalidity ratio, which is lowest for GPT-6
  Astra.
  Among the task-specific methods it also generates valid models most reliably, with an invalidity ratio
  of $0.67\%$ against $0.74\%$ for the closest baseline.
  Its outputs agree with the references in structure rather than in shape alone, as reflected by the
  highest intersection over union and Euler characteristic match, and its mean Chamfer distance is $1.14$
  against $1.75$ for the strongest general-purpose model.

  As shown in Figure~\ref{fig:results}(a,b), the error distribution of CAD-WAM dominates those of the
  task-specific baselines over the entire range, and the difference is largest at the upper end, where the
  median Chamfer distance decreases by $15\%$ and the mean by $56\%$ relative to the strongest baseline on each metric.
  Grouped by construction length (Figure~\ref{fig:results}(c,d)), CAD-WAM attains the lowest mean error in
  every group, and the gap to the baselines widens as models require more actions.
  The invalidity ratio of CAD-WAM remains low, whereas that of the baselines grows with the
  number of actions.
\subsection{Latent versus Textual Geometric States}
  \label{sec:latent_vs_text}
\begin{table}[t]
  \centering
  \caption{
  Comparison between latent geometric states and textual states.
  }
  \label{tab:latent_vs_text_state}

  \small
  \setlength{\tabcolsep}{3pt}
  \renewcommand{\arraystretch}{1.15}

  \resizebox{\linewidth}{!}{
  \begin{tabular}{
      @{}
      c
      c
      cc
      cc
      cc
      c
      c
      c
      @{}
  }
      \toprule

      \multirow{2}{*}{\textbf{State}}
      & \multirow{2}{*}{\textbf{IR (\%) $\downarrow$}}
      & \multicolumn{2}{c}{\textbf{CD $\downarrow$}}
      & \multicolumn{2}{c}{\textbf{HD $\downarrow$}}
      & \multicolumn{2}{c}{\textbf{NC $\uparrow$}}
      & \multirow{2}{*}{\textbf{ECD $\downarrow$}}
      & \multirow{2}{*}{\textbf{IoU $\uparrow$}}
      & \multirow{2}{*}{\textbf{Euler match (\%) $\uparrow$}}
      \\

      \cmidrule(lr){3-4}
      \cmidrule(lr){5-6}
      \cmidrule(lr){7-8}

      &
      & \textit{mean}
      & \textit{median}
      & \textit{mean}
      & \textit{median}
      & \textit{mean}
      & \textit{median}
      &
      &
      &
      \\

      \midrule

      \textbf{\shortstack[c]{Latent State}}
      & \textbf{0.666}
      & \textbf{1.142}
      & \textbf{0.166}
      & \textbf{4.134}
      & \textbf{2.783}
      & \textbf{0.903}
      & \textbf{0.964}
      & \textbf{7.313}
      & \textbf{0.964}
      & \textbf{95.850}
      \\

      \textbf{Text State}
      & 1.482
      & 1.405
      & 0.172
      & 4.619
      & 2.834
      & 0.893
      & 0.964
      & 7.996
      & 0.952
      & 94.556
      \\

      \bottomrule
  \end{tabular}
  }

  \end{table}
To examine the role of the state representation, we train a variant in which the geometric state is
  expressed in text rather than encoded into a latent vector, and in which the next state is likewise
  generated as text (Appendix~\ref{app:text_state}).
  This variant therefore requires neither the Geometric State Embedding nor the State Prediction Head, and
  all remaining factors are identical.

  As shown in Figure~\ref{fig:results}(a,b), the error distribution of CAD-WAM lies above those of the
    task-specific baselines over the entire range, and its quartile box and whiskers are the shortest of
    all systems; the difference is largest in the upper tail, where the median Chamfer distance decreases
    by $15\%$ and the mean by $56\%$ relative to the strongest baseline on each measure.
    Grouped by construction length (Figure~\ref{fig:results}(c,d)), CAD-WAM attains the lowest mean
    Chamfer distance in every group, and the gap widens as models require more actions.
    The invalidity ratio of CAD-WAM remains low across all groups, whereas that of the task-specific
    baselines grows with the number of actions.
\begin{table}[t]
  \centering
  \caption{
  Ablation study on the State Prediction Head.
  }
  \label{tab:state_prediction_head_ablation}

  \small
  \setlength{\tabcolsep}{3pt}
  \renewcommand{\arraystretch}{1.15}

  \resizebox{\linewidth}{!}{
  \begin{tabular}{
      @{}
      l
      c
      cc
      cc
      cc
      c
      c
      c
      @{}
  }
      \toprule

      \multirow{2}{*}{\textbf{Configuration}}
      & \multirow{2}{*}{\textbf{IR (\%) $\downarrow$}}
      & \multicolumn{2}{c}{\textbf{CD $\downarrow$}}
      & \multicolumn{2}{c}{\textbf{HD $\downarrow$}}
      & \multicolumn{2}{c}{\textbf{NC $\uparrow$}}
      & \multirow{2}{*}{\textbf{ECD $\downarrow$}}
      & \multirow{2}{*}{\textbf{IoU $\uparrow$}}
      & \multirow{2}{*}{\textbf{Euler match (\%) $\uparrow$}}
      \\

      \cmidrule(lr){3-4}
      \cmidrule(lr){5-6}
      \cmidrule(lr){7-8}

      &
      & \textit{mean}
      & \textit{median}
      & \textit{mean}
      & \textit{median}
      & \textit{mean}
      & \textit{median}
      &
      &
      &
      \\

      \midrule

      w/o State Prediction Head
      & 0.779
      & 1.316
      & 0.167
      & 4.264
      & 2.785
      & 0.902
      & \textbf{0.964}
      & 7.374
      & 0.960
      & 95.404
      \\

      w/ State Prediction Head
      & \textbf{0.666}
      & \textbf{1.142}
      & \textbf{0.166}
      & \textbf{4.134}
      & \textbf{2.783}
      & \textbf{0.903}
      & \textbf{0.964}
      & \textbf{7.313}
      & \textbf{0.964}
      & \textbf{95.850}
      \\

      \bottomrule
  \end{tabular}
  }

  \end{table}
\subsection{Ablation on the State Prediction Head}
\label{sec:ablation}

  To assess the contribution of state prediction, we train a variant in which the State Prediction Head
  is removed, so that the next geometric state is no longer predicted.
  All remaining settings are unchanged, so the two configurations differ only in the presence of this
  head.

  As reported in Table~\ref{tab:state_prediction_head_ablation}, state prediction improves every
  aggregate metric.
  The invalidity ratio decreases from $0.78\%$ to $0.67\%$ and the mean Chamfer distance from $1.32$ to
  $1.14$, whereas the medians of the two configurations coincide.
  The improvement is therefore concentrated in the tail of the error distribution and is accompanied by a
  higher intersection over union and a higher Euler characteristic match. These results indicate that predicting the geometric state that should result from the next action,
  rather than generating the action alone, leads to more reliable generation.

\section{Conclusion}
We presented CAD-WAM, a text-to-CAD framework that couples geometry-conditioned action generation with next-state prediction. Its operation-level DSL and latent geometric states allow each modeling step to be learned through both an executable action and its intended geometric outcome. Next-state supervision shapes the shared representation used for action generation, while CAD-kernel execution provides feedback from the geometry actually built. On the DeepCAD test set, CAD-WAM achieves higher geometric accuracy than the evaluated task-specific and general-purpose models, with larger gains on models requiring more operations. Ablations show that the latent-state model outperforms its textual-state counterpart and that next-state supervision further improves generation. Together, these results support learning geometric evolution alongside command generation as an effective approach to text-to-CAD modeling.

\subsection*{AI use statement}
In this work, we used generative AI tools to aid and polish the writing of the paper, including grammatical correction and rephrasing for clarity and concision. We have reviewed all AI-assisted work: every assisted passage was read and, where necessary, rewritten by the authors, and all numbers and claims reported in the paper were checked against our own experimental records. We take responsibility for the final content of this work, including text, claims and artifacts.

\subsection*{Reproducibility statement}

  The results of this paper rest on three components, each documented: the action
  representation together with the geometric executor that defines every state transition, the data
  construction and training procedure that learns from those states, and a single evaluation protocol
  applied uniformly to our model and to every baseline. The main paper states the model formulation and
  the experimental setting, while the Appendix specifies the metric definitions, the construction of the
  evaluation set, and the remaining implementation and ablation details. Since the same executor produces
  the states supervised during training and those fed back during closed-loop generation, our results can
  be reproduced end to end rather than only at the level of the final geometry. 
  
\bibliography{references}

@inproceedings{khan2024text2cad,
 author = {Khan, Mohammad Sadil and Sinha, Sankalp and Sheikh, Talha Uddin and Stricker, Didier and Ali, Sk Aziz and Afzal, Muhammad Zeshan},
 booktitle = {Advances in Neural Information Processing Systems},
 doi = {10.52202/079017-0242},
 pages = {7552--7579},
 publisher = {Curran Associates, Inc.},
 title = {{{Text2CAD}: Generating Sequential {CAD} Designs from Beginner-to-Expert Level Text Prompts}},
 volume = {37},
 year = {2024}
}

@inproceedings{guan2025cadcoder,
 author = {Guan, Yandong and Wang, Xilin and Xing, Ximing and Zhang, Jing and Xu, Dong and Yu, Qian},
 booktitle = {Advances in Neural Information Processing Systems},
 doi = {10.52202/085713-1999},
 pages = {59765--59789},
 publisher = {Curran Associates, Inc.},
 title = {{{CAD-Coder}: Text-to-{CAD} Generation with Chain-of-Thought and Geometric Reward}},
 volume = {38},
 year = {2025}
}

@inproceedings{pyatov2026cadfs,
 author = {Pyatov, Vladislav and Bobrovskikh, Gleb and Galochkin, Saveliy and Boldyrev, Nikita and Voynov, Oleg and Filippov, Alexander and Ferrer, Gonzalo and Wonka, Peter and Burnaev, Evgeny},
 booktitle = {Proceedings of the IEEE/CVF Conference on Computer Vision and Pattern Recognition (CVPR)},
 pages = {10176--10186},
 title = {{CADFS: A Big CAD Program Dataset and Framework for Computer-Aided Design with Large Language Models}},
 year = {2026}
}

@article{kolodiazhnyi2026cadrille,
 archiveprefix = {arXiv},
 author = {Kolodiazhnyi, Maksim and Tarasov, Denis and Zhemchuzhnikov, Dmitrii and Nikulin, Alexander and Zisman, Ilya and Vorontsova, Anna and Konushin, Anton and Kurenkov, Vladislav and Rukhovich, Danila},
 eprint = {2505.22914v3},
 journal = {arXiv preprint arXiv:2505.22914v3},
 note = {Version 3, 17 February 2026},
 primaryclass = {cs.CV},
 title = {{cadrille: Multi-modal CAD Reconstruction with Reinforcement Learning}},
 year = {2026}
}

@inproceedings{qi2026pointercad,
 author = {Qi, Dacheng and Wang, Chenyu and Xu, Jingwei and Chu, Tianzhe and Zhao, Zibo and Liu, Wen and Ding, Wenrui and Ma, Yi and Gao, Shenghua},
 booktitle = {Proceedings of the IEEE/CVF Conference on Computer Vision and Pattern Recognition (CVPR)},
 pages = {17377--17387},
 title = {{Pointer-CAD: Unifying B-Rep and Command Sequences via Pointer-based Edges \& Faces Selection}},
 year = {2026}
}

@article{ha2018world,
 author = {Ha, David and Schmidhuber, J{\"u}rgen},
 doi = {10.48550/arXiv.1803.10122},
 journal = {arXiv preprint arXiv:1803.10122},
 title = {{World Models}},
 year = {2018}
}

@inproceedings{hafner2020dream,
 author = {Hafner, Danijar and Lillicrap, Timothy and Ba, Jimmy and Norouzi, Mohammad},
 booktitle = {International Conference on Learning Representations},
 title = {{Dream to Control: Learning Behaviors by Latent Imagination}},
 year = {2020}
}

@inproceedings{tian2025seer,
 author = {Tian, Yang and Yang, Sizhe and Zeng, Jia and Wang, Ping and Lin, Dahua and Dong, Hao and Pang, Jiangmiao},
 booktitle = {International Conference on Learning Representations},
 title = {{Predictive Inverse Dynamics Models are Scalable Learners for Robotic Manipulation}},
 year = {2025}
}

@inproceedings{zhu2025uwm,
 author = {Zhu, Chuning and Yu, Raymond and Feng, Siyuan and Burchfiel, Benjamin and Shah, Paarth and Gupta, Abhishek},
 booktitle = {Proceedings of Robotics: Science and Systems},
 doi = {10.15607/RSS.2025.XXI.015},
 title = {{Unified World Models: Coupling Video and Action Diffusion for Pretraining on Large Robotic Datasets}},
 year = {2025}
}

@article{ye2026gigaworld,
 author = {Ye, Angen and Wang, Boyuan and Ni, Chaojun and Huang, Guan and Zhao, Guosheng and Li, Hao and Li, Hengtao and Li, Jie and Lv, Jindi and Liu, Jingyu and Cao, Min and Li, Peng and Deng, Qiuping and Mei, Wenjun and Wang, Xiaofeng and Chen, Xinze and Zhou, Xinyu and Wang, Yang and Chang, Yifan and Li, Yifan and Zhou, Yukun and Ye, Yun and Liu, Zhichao and Zhu, Zheng},
 doi = {10.48550/arXiv.2603.17240},
 journal = {arXiv preprint arXiv:2603.17240},
 title = {{GigaWorld-Policy: An Efficient Action-Centered World--Action Model}},
 year = {2026}
}

@inproceedings{kim2021setvae,
 author = {Kim, Jinwoo and Yoo, Jaehoon and Lee, Juho and Hong, Seunghoon},
 booktitle = {Proceedings of the IEEE/CVF Conference on Computer Vision and Pattern Recognition (CVPR)},
 pages = {15059--15068},
 title = {{SetVAE: Learning Hierarchical Composition for Generative Modeling of Set-Structured Data}},
 year = {2021}
}

@inproceedings{wu2021deepcad,
 author = {Wu, Rundi and Xiao, Chang and Zheng, Changxi},
 booktitle = {Proceedings of the IEEE/CVF International Conference on Computer Vision (ICCV)},
 pages = {6772--6782},
 title = {{DeepCAD: A Deep Generative Network for Computer-Aided Design Models}},
 year = {2021}
}

@inproceedings{xu2022skexgen,
 author = {Xu, Xiang and Willis, Karl D. D. and Lambourne, Joseph G. and Cheng, Chin-Yi and Jayaraman, Pradeep Kumar and Furukawa, Yasutaka},
 booktitle = {Proceedings of the 39th International Conference on Machine Learning},
 pages = {24698--24724},
 publisher = {PMLR},
 series = {Proceedings of Machine Learning Research},
 title = {{SkexGen: Autoregressive Generation of CAD Construction Sequences with Disentangled Codebooks}},
 volume = {162},
 year = {2022}
}

@inproceedings{xu2023hnccad,
 author = {Xu, Xiang and Jayaraman, Pradeep Kumar and Lambourne, Joseph George and Willis, Karl D. D. and Furukawa, Yasutaka},
 booktitle = {Proceedings of the 40th International Conference on Machine Learning},
 pages = {38443--38461},
 publisher = {PMLR},
 series = {Proceedings of Machine Learning Research},
 title = {{Hierarchical Neural Coding for Controllable CAD Model Generation}},
 volume = {202},
 year = {2023}
}

@inproceedings{rukhovich2025cadrecode,
 author = {Rukhovich, Danila and Dupont, Elona and Mallis, Dimitrios and Cherenkova, Kseniya and Kacem, Anis and Aouada, Djamila},
 booktitle = {Proceedings of the IEEE/CVF International Conference on Computer Vision (ICCV)},
 doi = {10.1109/ICCV51701.2025.00914},
 pages = {9801--9811},
 title = {{CAD-Recode: Reverse Engineering CAD Code from Point Clouds}},
 year = {2025}
}

@article{govindarajan2025cadmium,
 author = {Govindarajan, Prashant and Baldelli, Davide and Pathak, Jay and Fournier, Quentin and Chandar, Sarath},
 journal = {Transactions on Machine Learning Research},
 title = {{{CADmium}: Fine-Tuning Code Language Models for Text-Driven Sequential {CAD} Design}},
 url = {https://openreview.net/forum?id=lExqWvQht8},
 year = {2026}
}

@inproceedings{du2023unipi,
 author = {Du, Yilun and Yang, Mengjiao and Dai, Bo and Dai, Hanjun and Nachum, Ofir and Tenenbaum, Joshua B. and Schuurmans, Dale and Abbeel, Pieter},
 booktitle = {Advances in Neural Information Processing Systems},
 doi = {10.52202/075280-0403},
 pages = {9156--9172},
 publisher = {Curran Associates, Inc.},
 title = {{Learning Universal Policies via Text-Guided Video Generation}},
 volume = {36},
 year = {2023}
}

@inproceedings{bruce2024genie,
 author = {Bruce, Jake and Dennis, Michael D and Edwards, Ashley and Parker-Holder, Jack and Shi, Yuge and Hughes, Edward and Lai, Matthew and Mavalankar, Aditi and Steigerwald, Richie and Apps, Chris and Aytar, Yusuf and Bechtle, Sarah Maria Elisabeth and Behbahani, Feryal and Chan, Stephanie C.Y. and Heess, Nicolas and Gonzalez, Lucy and Osindero, Simon and Ozair, Sherjil and Reed, Scott and Zhang, Jingwei and Zolna, Konrad and Clune, Jeff and Freitas, Nando De and Singh, Satinder and Rockt{\"a}schel, Tim},
 booktitle = {Proceedings of the 41st International Conference on Machine Learning},
 pages = {4603--4623},
 publisher = {PMLR},
 series = {Proceedings of Machine Learning Research},
 title = {{Genie: Generative Interactive Environments}},
 volume = {235},
 year = {2024}
}

@misc{cadquery,
 author = {{CadQuery contributors}},
 doi = {10.5281/zenodo.18633916},
 howpublished = {Computer software, version 2.7.0. Zenodo},
 title = {{CadQuery}},
 version = {2.7.0},
 year = {2026}
}

@inproceedings{alayrac2022flamingo,
 author = {Alayrac, Jean-Baptiste and Donahue, Jeff and Luc, Pauline and Miech, Antoine and Barr, Iain and Hasson, Yana and Lenc, Karel and Mensch, Arthur and Millican, Katherine and Reynolds, Malcolm and Ring, Roman and Rutherford, Eliza and Cabi, Serkan and Han, Tengda and Gong, Zhitao and Samangooei, Sina and Monteiro, Marianne and Menick, Jacob L and Borgeaud, Sebastian and Brock, Andy and Nematzadeh, Aida and Sharifzadeh, Sahand and Bi\'{n}kowski, Miko\l aj and Barreira, Ricardo and Vinyals, Oriol and Zisserman, Andrew and Simonyan, Kar\'{e}n},
 booktitle = {Advances in Neural Information Processing Systems},
 doi = {10.52202/068431-1723},
 pages = {23716--23736},
 publisher = {Curran Associates, Inc.},
 title = {{Flamingo: a Visual Language Model for Few-Shot Learning}},
 volume = {35},
 year = {2022}
}

@inproceedings{li2023blip2,
 author = {Li, Junnan and Li, Dongxu and Savarese, Silvio and Hoi, Steven},
 booktitle = {Proceedings of the 40th International Conference on Machine Learning},
 pages = {19730--19742},
 publisher = {PMLR},
 series = {Proceedings of Machine Learning Research},
 title = {{{BLIP-2}: Bootstrapping Language-Image Pre-training with Frozen Image Encoders and Large Language Models}},
 volume = {202},
 year = {2023}
}

@misc{qwen2026qwen38flash,
 author = {{Alibaba Cloud}},
 howpublished = {Alibaba Cloud Model Studio product announcement},
 month = {August},
 note = {Published 2026-08-28; accessed 2026-09-26},
 title = {{{Qwen3.8-Flash} Released}},
 url = {https://www.alibabacloud.com/en/news/product/qwen38-flash-released-noi},
 year = {2026}
}

@misc{deepseek2026v41flash,
 author = {{DeepSeek-AI}},
 howpublished = {Technical report},
 note = {Accessed 2026-09-26},
 title = {{{DeepSeek-V4.1-Flash}: Pushing the Limits of {KV} Cache Compression}},
 url = {https://huggingface.co/deepseek-ai/DeepSeek-V4.1-Flash},
 year = {2026}
}

@misc{openai2026gpt6astra,
 author = {{OpenAI}},
 howpublished = {System card},
 note = {Accessed 2026-09-26},
 title = {{{GPT-6 Astra} System Card}},
 url = {https://deploymentsafety.openai.com/gpt-6-astra},
 year = {2026}
}

@misc{anthropic2026opus55,
 author = {{Anthropic}},
 howpublished = {System card},
 note = {Accessed 2026-09-26},
 title = {{{Claude Opus 5.5} System Card}},
 url = {https://www.anthropic.com/claude-opus-5-5-system-card},
 year = {2026}
}

@article{qwen2024qwen25,
 author = {{Qwen Team}},
 doi = {10.48550/arXiv.2412.15115},
 journal = {arXiv preprint arXiv:2412.15115},
 title = {{{Qwen2.5} Technical Report}},
 year = {2024}
}
\bibliographystyle{iclr2027_conference}
  
\newpage 
\appendix
  \section*{Appendix}

\section{CAD Domain-Specific Language}
  \label{app:dsl}

  We detail here the action format introduced in Section~\ref{sec:dsl} and the checks applied before and
  after execution.

  \subsection{Action Schema}

  An action is a JSON object identified by its \texttt{kind} field; the extrusion used throughout our
  dataset is specified in Table~\ref{tab:dsl_fields}, and Figure~\ref{fig:dsl} shows such an action in its
  serialized form.
  Its sketch plane is given by an origin and two orthonormal directions, which define the local
  two-dimensional frame in which the profile is expressed.
  A profile consists of one or more closed loops, each a chain of curves in which every curve begins at
  the end point of its predecessor and the final curve closes the loop.
  Loops nested inside other loops define holes according to the even-odd rule.
  Lines are specified by their end point, arcs additionally by a sweep angle and an orientation flag, and
  circles by a center and a radius; a circle is always the only curve of its loop.

  \begin{table}[ht]
  \centering
  \caption{Fields of an extrusion action.}
  \label{tab:dsl_fields}
  \small
  \begin{tabular}{@{}lll@{}}
  \toprule
  Field & Type & Meaning \\
  \midrule
  \texttt{kind} & string & operation identifier, \texttt{"extrude"} \\
  \texttt{plane.origin} & 3 floats & origin of the sketch plane \\
  \texttt{plane.normal} & 3 floats & plane normal, along which the profile is extruded \\
  \texttt{plane.x\_dir} & 3 floats & in-plane $x$ direction \\
  \texttt{profile.loops} & list & closed loops of lines, arcs and circles \\
  \texttt{forward} & float $\geq 0$ & extrusion distance along the normal \\
  \texttt{backward} & float $\geq 0$ & extrusion distance against the normal \\
  \texttt{op} & enum & \texttt{new}, \texttt{join}, \texttt{cut} or \texttt{intersect} \\
  \texttt{taper} & float & draft angle in degrees, zero throughout DeepCAD \\
  \bottomrule
  \end{tabular}
  \end{table}

  The pair $(\texttt{forward}, \texttt{backward})$ covers the three extrusion types of the source data:
  one of the two vanishes for a one-sided extrusion, both are equal for a symmetric one, and both are
  positive and distinct for a two-sided one.
  The Boolean operation then combines the extruded body with the current model.
  Because our executor fuses a new body with the existing model, \texttt{new} and \texttt{join} have the
  same geometric effect, and a \texttt{join} applied to an empty model is treated as \texttt{new}.
  A construction terminates with the dedicated action \texttt{\{"kind": "end"\}}.Figure~\ref{fig:dsl_example} shows a complete action in its serialized form.

  \begin{figure}[ht]
  \centering
  \begin{tcblisting}{listing only,
    colback=white, colframe=black!55, boxrule=0.5pt, arc=2pt,
    left=6pt, right=6pt, top=4pt, bottom=4pt, width=\linewidth,
    title=\small An action of our CAD DSL,
    listing options={basicstyle=\ttfamily\footnotesize, columns=flexible, breaklines=true}}
  {
    "kind": "extrude",
    "plane": {
      "origin": [0, 0, 0],
      "normal": [0, 0, 1],
      "x_dir":  [1, 0, 0]
    },
    "profile": {
      "loops": [
        {"curves": [
          {"type": "line", "end": [0.0254, 0]},
          {"type": "arc",  "end": [0, -0.0254],
           "sweep": 1.5708, "ccw": false},
          {"type": "line", "end": [0, -0.254]},
          {"type": "line", "end": [0.254, -0.254]},
          {"type": "line", "end": [0.254, 0]}
        ]}
      ]
    },
    "forward": 0.0127,
    "backward": 0,
    "op": "new",
    "taper": 0
  }
  \end{tcblisting}
  \caption{An action of our DSL. A square profile with one rounded corner is drawn on the $xy$-plane and
  extruded as a new body.}
  \label{fig:dsl_example}
  \end{figure}

  \subsection{Execution and Validation}

  Every action is validated before it reaches the CAD kernel.
  Beyond conformance to the schema, we require non-empty loops, a positive radius for a circle that forms
  its own loop, curves of non-zero length relative to the size of their loop, arc sweeps strictly between
  $0$ and $2\pi$, loops of non-zero area, a positive total extrusion distance, and a Boolean operation
  applicable to the current model, since \texttt{cut} and \texttt{intersect} are undefined on an empty
  model.

  Actions that pass these checks are executed by the OpenCASCADE kernel through CadQuery~\citep{cadquery}.
  A result is accepted only if it comprises at least one solid, passes the kernel's validity check, and
  has a volume exceeding $10^{-9}$ times the cube of its bounding-box diagonal.
  We additionally require the volume change to agree with the Boolean operation, within a relative
  tolerance of $10^{-6}$: a \texttt{join} may not decrease the volume, and a \texttt{cut} or an
  \texttt{intersect} may not increase it.
  Violations trigger a single retry with a fuzzy Boolean tolerance of $10^{-6}$ times the bounding-box
  diagonal, after which the action is rejected if the result remains inconsistent.
  A rejected action leaves the current model unchanged, so that every accepted action yields a valid solid.
  \section{SetVAE Pretraining}
  \label{app:setvae}

  The encoder $E_\phi$ of Section~\ref{sec:state} is obtained by training SetVAE on the point clouds of
  the intermediate and final solids of the training split, that is 317{,}711 clouds from 157{,}945 models.
  A further 17{,}639 clouds of the validation split are used for checkpoint selection, and the clouds of
  the test split are never observed during this stage.
  Each cloud contains 2{,}048 points sampled over the surface of a solid, centered at its bounding box and
  scaled to a maximum half-extent of $0.75$, after which the training-set mean and standard deviation are
  applied, following the original implementation.

  The model follows the SetVAE architecture~\citep{kim2021setvae} with seven latent levels holding 1, 1,
  2, 4, 8, 16 and 32 latent vectors of dimension 16, so that the concatenated posterior means form the
  $1024$-dimensional geometric state.
  Training minimizes a Chamfer reconstruction term together with the Kullback-Leibler divergence, whose
  weight is annealed from zero to one over the first 40 epochs.
  We use Adam with a learning rate of $10^{-3}$, a warm-up of 1{,}400 steps followed by a cosine decay to
  one percent of that value, a global batch of 256 clouds, and bf16 precision, for 300 epochs on eight
  H800 GPUs, which takes approximately 11.5 hours.
  Validation is performed every five epochs, and the checkpoint with the lowest validation Chamfer
  distance is retained; it occurs at the final epoch, with a validation Chamfer distance of $39.02$
  against $38.28$ on the training clouds.

\begin{figure}[t]
  \centering
  \includegraphics[width=\linewidth]{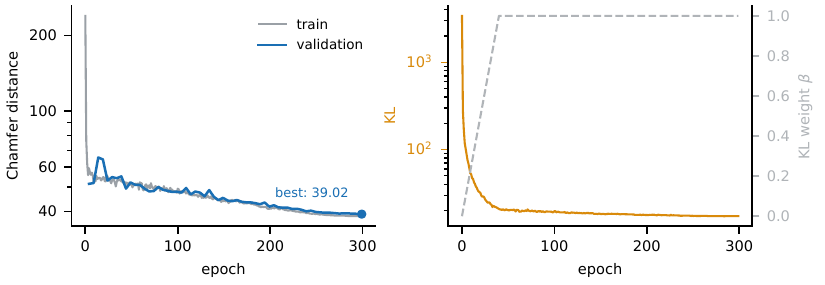}
  \caption{SetVAE pretraining. Left: training and validation Chamfer distance per epoch on a logarithmic
  scale, with the selected checkpoint marked. Right: the Kullback-Leibler term and its weight schedule.}
  \label{fig:setvae_curves}
  \end{figure}

\begin{figure}[t]
  \centering
  \includegraphics[width=\linewidth]{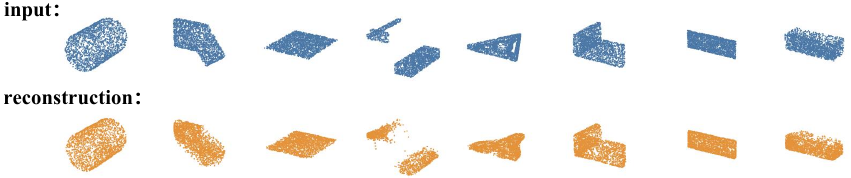}
  \caption{Reconstructions of test-split solids that the encoder never observed during pretraining,
  including intermediate solids from several construction steps. Input clouds are shown above their
  reconstructions.}
  \label{fig:setvae_recon}
  \end{figure}

  Figure~\ref{fig:setvae_curves} shows the training and validation curves.
  The transient increase of the validation error around the twentieth epoch coincides with the annealing
  of the Kullback-Leibler term, and the two curves remain close throughout, indicating that the model does
  not overfit the training clouds.
  Figure~\ref{fig:setvae_recon} compares input clouds of the test split with their reconstructions.
  Over 512 randomly selected test clouds, the reconstruction Chamfer distance is $2.44\times10^{-3}$ on
  average, with a median of $1.96\times10^{-3}$ and a 90th percentile of $4.74\times10^{-3}$, measured in
  the normalized frame in which each solid has a maximum half-extent of $0.75$.
  Planar faces, holes and overall proportions are preserved, whereas fine edges are slightly rounded,
  which delimits the geometric detail that the state can represent.
  After this stage, $E_\phi$ is kept frozen and is used only to encode point clouds into geometric states.

  \section{Implementation Details}
  \label{app:impl}

  Table~\ref{tab:impl} lists the configuration of CAD-WAM. All models are trained and evaluated on eight
  NVIDIA H800 GPUs with 80\,GB of memory.

  \begin{table}[t]
  \centering
  \caption{Configuration of CAD-WAM.}
  \label{tab:impl}
  \small
  \begin{tabular}{@{}ll@{}}
  \toprule
  \multicolumn{2}{@{}l}{\textit{Architecture}} \\
  Backbone & Qwen2.5-1.5B-Instruct \\
  Geometric tokens & 16 \\
  Slot width & 256 \\
  Geometric State Embedding & 2 Transformer layers, 8 heads, feed-forward width 1024 \\
  State Prediction Head & mean pooling, LayerNorm, MLP $d \rightarrow 2048 \rightarrow 1024$ (GELU) \\
  Added parameters & 11.4M, or 0.74\% of the backbone \\
  \midrule
  \multicolumn{2}{@{}l}{\textit{Training}} \\
  Hardware & 8 $\times$ NVIDIA H800 \\
  Precision and parallelism & bf16, fully sharded data parallelism \\
  Epochs & 1 \\
  Records per optimizer step & 64, from micro-batch 4 $\times$ 8 GPUs $\times$ 2 accumulation \\
  Optimizer steps & $\approx 7.4$k \\
  Learning rate & $2\times10^{-5}$ for the backbone, $1\times10^{-4}$ for the new modules \\
  Schedule & 3\% warm-up, cosine decay to 10\% of the peak \\
  Gradient clipping & 1.0 \\
  Maximum sequence length & 5{,}632 \\
  State loss weight $\lambda$ & 0.03 \\
  \midrule
  \multicolumn{2}{@{}l}{\textit{Inference}} \\
  Decoding & greedy \\
  Maximum tokens per action & 2{,}048 \\
  Maximum number of actions & 20 \\
  Points per solid & 4{,}096 sampled, fixed subset of 2{,}048 encoded \\
  \bottomrule
  \end{tabular}
  \end{table}

  \paragraph{Fusion of the geometric tokens.}
  The tokenized prompt contains sixteen contiguous reserved positions.
  Their token embeddings are replaced by the output of the Geometric State Embedding before the sequence
  enters the backbone.
  No layer of the decoder stack is modified.
  Gradients are not detached between the two branches, so both objectives update the backbone and the
  Geometric State Embedding.

  \paragraph{Optimization.}
  The cross-entropy is computed over the action tokens only, and the prompt is masked out.
  The state term is evaluated in single precision, independently of the compute precision.
  Fully sharded data parallelism wraps the model at the granularity of a decoder layer.
  All geometric states are precomputed and stored in a memory-mapped array, so each training step reads
  them without recomputation.
  The dataset is built so that no prompt exceeds the maximum sequence length.
  We validate on a fixed random subset of the validation split and report the cross-entropy of action
  tokens and state tokens separately.

  \paragraph{Inference.}
  Each action is decoded greedily and passed to a geometry worker.
  The workers run in a separate process and in the environment that produced the training point clouds,
  so the encoder observes the same tessellation at training and at test time.
  They are stateless, which isolates kernel failures from the generation process.
  The solid returned by a worker is sampled, normalized as in Appendix~\ref{app:setvae}, and encoded into
  the next geometric state.
  Test models are distributed over the eight GPUs, and the geometry workers of each shard run in parallel
  with generation.

  \paragraph{Software environment.}
    Training and generation use Python 3.12 with PyTorch 2.8.0 built against CUDA 12.8,
    Transformers 4.57.1 and FlashAttention 2.8.1, on GPUs with driver version 590.48.01.
    Generation and evaluation are kept in separate environments: the geometry workers run the kernel
    version that produced the training data, whereas the evaluation environment applies one kernel version
    to the outputs of all systems.
 \section{Dataset Construction}
  \label{app:data}

  \paragraph{Source data.}
  The CAD models and their construction sequences come from DeepCAD~\citep{wu2021deepcad}, which provides
  178{,}238 models together with a reference solid for each of them.
  The textual targets are the expert-level descriptions provided by Recap-OmniCAD~\citep{qi2026pointercad}.
  We keep DeepCAD's official split and do not remove duplicate programs.

  \paragraph{Conversion to the DSL.}
  Every sketch profile and its extrusion are converted into one action of our DSL.
  The sketch plane is taken from the transform of the sketch, and the loops of the profile are taken from
  its curves.
  The two extrusion distances follow the extent type of the feature, which is one-sided, symmetric or
  two-sided.
  The Boolean operation follows the feature operation of the original model.
  An extrusion that references several profiles is split into a sequence of single-profile actions, which
  leaves the resulting geometry unchanged for the operations that occur in DeepCAD.
  All coordinates are kept in the units of the source data, and no rescaling is applied.

  \paragraph{Verification.}
  Each converted program is executed and the resulting solid is compared with the reference solid.
  A model is retained when the maximum surface distance between the two solids is below $10^{-6}$.
  This procedure verifies 176{,}226 of the 178{,}238 models, and 174{,}582 of those also carry a
  description.
  The remaining models fail because the conversion does not complete, because the executor rejects an
  action, or because the resulting geometry differs from the reference.
  Table~\ref{tab:data_census} summarizes the construction of the dataset.

  \begin{table}[ht]
  \centering
  \caption{Construction of the step-wise dataset. The test split lists the models for which step-wise
  records exist, whereas the evaluation set comprises all test models that carry a description,
  including 80 models that do not pass verification.}
  \label{tab:data_census}
  \small
  \begin{tabular}{@{}lrr@{}}
  \toprule
   & Models & Records \\
  \midrule
  DeepCAD models & 178{,}238 & \\
  Geometrically verified & 176{,}226 & \\
  With an expert description & 174{,}582 & 526{,}010 \\
  \midrule
  Training split & 157{,}945 & 475{,}656 \\
  Validation split & 8{,}755 & 26{,}394 \\
  Test split & 7{,}882 & 23{,}960 \\
  Evaluation set & 7{,}962 & \\
  \bottomrule
  \end{tabular}
  \end{table}

  \paragraph{Step-wise records.}
  Each program is executed one action at a time, which yields the intermediate solids of the construction.
  Every intermediate solid is sampled and encoded into a geometric state with the pretrained encoder of
  Appendix~\ref{app:setvae}.
  A record is then formed for every step, and a final record supervises the end action.
  Of the 526{,}010 records, 351{,}428 correspond to a modeling action and the remainder to the end action.
  The records are stored in columnar files, and the geometric states are stored in a memory-mapped array
  whose first row holds the zero state of the empty model.
  A record can only be built for a model whose program executes and matches its reference, which is why
  the test split of the dataset is limited to the verified models.
  Generation, in contrast, starts from the empty model and requires only the description.
  We therefore evaluate on every test model that carries a description, so that the evaluation is not
  restricted to the models our conversion reproduces.

\begin{figure}[t]
  \centering
  \includegraphics[width=\linewidth]{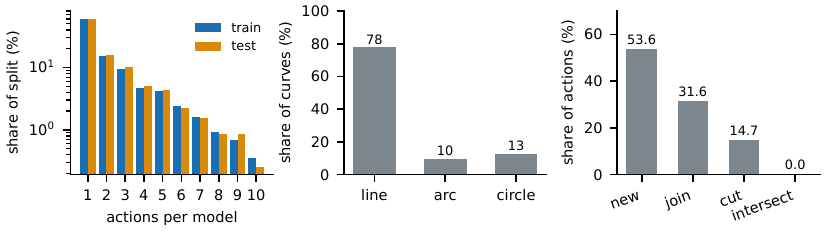}
  \caption{Statistics of the step-wise dataset. Left: number of actions per model, shown as a share of
  the training and of the test split on a logarithmic scale. Middle: composition of the sketch profiles
  by curve type. Right: distribution of the Boolean operations over all actions.}
  \label{fig:dataset_stats}
  \end{figure}

  \paragraph{Statistics.}
  Figure~\ref{fig:dataset_stats} summarizes the resulting programs.
  The number of actions per model ranges from one to ten, and the test split
  follows the same distribution as the training split.
  Sketch profiles consist of 1{,}139{,}814 lines, 184{,}224 circles and 142{,}333 arcs, and 279{,}304
  profiles contain a single loop.
  Among the Boolean operations, 188{,}528 create a new body, 111{,}139 join, 51{,}651 cut and 110
  intersect.
  Every action in DeepCAD uses a zero draft angle.
  Figure~\ref{fig:record_example} shows a record of the dataset.

  \begin{figure}[t]
  \centering
  \begin{tcolorbox}[colback=white, boxrule=0.5pt, left=4pt, right=4pt, top=3pt, bottom=3pt,
                    title=\small Input]
  \ttfamily\scriptsize
  You are constructing a CAD model step by step with cadseq version 1 actions. You are given
  the CAD description of the target model and the current geometric state of the partially
  built model, encoded as 16 geometric state tokens. Generate ONLY the next cadseq action as
  a single JSON object, preserving the described dimensions. If the model is already
  complete, return \{"kind": "end"\}.
  \\[4pt]
  CAD description:\\
  To construct the CAD model of the bracket, follow these detailed steps:\\
  \#\#\# Part 1: Symmetrical Square Bracket with Rounded Edges and Central Cutout\\
  1. Create Sketch 1 on the Top Plane: draw an outer loop of four arcs and four lines
  [\ldots] and an inner loop consisting of a single circle of radius 0.0025.\\
  2. Extrude the sketch to create a new body, extending 0.003 in the positive normal
  direction.\\
  \#\#\# Part 2: Central Circular Cutout with Smaller Concentric Ring [\ldots]
  \\[4pt]
  Actions completed so far: 1\\
  Current bounding box: size 0.02 x 0.02 x 0.003, center (0, 0, 0.0015)\\
  Current geometric state:\\
  <GEO>[16 geometric tokens]</GEO>
  \end{tcolorbox}

  \begin{tcolorbox}[colback=white, boxrule=0.5pt, left=4pt, right=4pt, top=3pt, bottom=3pt,
                    title=\small Target action]
  \ttfamily\scriptsize
  \{"kind": "extrude",\\
  \hspace*{1em}"plane": \{"origin": [0, 0, 0.003], "normal": [0, 0, 1],\\
  \hspace*{3.5em}"x\_dir": [1, 0, 0]\},\\
  \hspace*{1em}"profile": \{"loops": [\\
  \hspace*{2em}\{"curves": [\{"type": "circle", "center": [0, 0], "radius": 0.0025\}]\},\\
  \hspace*{2em}\{"curves": [\{"type": "circle", "center": [0, 0], "radius": 0.005\}]\}]\},\\
  \hspace*{1em}"forward": 0, "backward": 0.0015, "op": "cut", "taper": 0\}
  \end{tcolorbox}
  \caption{A record of the step-wise dataset, taken from the test split. The input contains the
  instruction, the description of the target model, the progress information and the geometric state of
  the current solid; the description is shortened here. The target is the next action, which cuts a ring
  into the bracket. The supervision also includes the geometric state of the solid that this action
  produces.}
  \label{fig:record_example}
  \end{figure}

  \section{Baseline Configurations}
  \label{app:baselines}

  We generate the outputs of Text2CAD, CAD-Coder and cadrille ourselves from the checkpoints released by
  their authors, and we use the released inference results of CADFS.
  For Pointer-CAD we report published numbers, since its generated models are not accessible to us.
  The general-purpose models are queried through their APIs with the step-wise protocol described below.
  Table~\ref{tab:baseline_config} summarizes the configurations.

 \begin{table}[ht]
  \centering
  \caption{Configuration of the compared systems. "Attempted" is the number of test models the system was
  run on.}
  \label{tab:baseline_config}
  \small
  \setlength{\tabcolsep}{6pt}
  \renewcommand{\arraystretch}{1.1}
  \begin{tabular}{@{}lccc@{}}
  \toprule
  System & Outputs & Generation & Attempted \\
  \midrule
  Text2CAD & our inference & single pass & 7{,}841 \\
  CAD-Coder & our inference & single pass & 7{,}958 \\
  cadrille & our inference & single pass & 7{,}949 \\
  CADFS & released results & single pass & 7{,}212 \\
  Pointer-CAD & published results & step-wise & -- \\
  \midrule
  Qwen3.8-Flash & API & DSL, rendered views & 7{,}962 \\
  DeepSeek-V4.1-Flash & API & DSL, rendered views & 7{,}962 \\
  GPT-6 Astra & API & DSL, rendered views & 7{,}962 \\
  Claude Opus 5.5 & API & DSL, rendered views & 7{,}962 \\
  \midrule
  CAD-WAM & our model & DSL, latent state & 7{,}962 \\
  \bottomrule
  \end{tabular}
  \end{table}

  \paragraph{Text2CAD.}
  We use the official checkpoint and the expert-level prompts released with the model, and decode greedily.
  Prompts are unavailable for 121 test models, which are therefore not attempted.
  Sequences are converted to solids with the post-processing used in the authors' evaluation script.
  Under our protocol this configuration reproduces the metrics reported in the original paper, which
  confirms that the pipeline is faithful.

  \paragraph{CAD-Coder.}
  We use the official checkpoint and the released test file, with the sampling settings of the authors:
  temperature 0.7, nucleus sampling with $p=0.7$, a limit of 4{,}096 tokens, and a fixed seed.
  No generation is truncated, and 99.5\% of the generated programs execute.

  \paragraph{cadrille.}
  We use the released checkpoint in its text-conditioned mode with the decoding settings of the authors.

  \paragraph{CADFS.}
  We evaluate the released inference results of the Qwen3-8B variant reported in the ablation study of the
  original paper.
  Of the attempted models, 586 produce no solid and are counted as invalid.

  \paragraph{Pointer-CAD.}
  The published results are obtained on the same test models and with the same Chamfer distance
  convention as ours, which makes them comparable to the re-scored systems.
  Metrics that the paper does not report are left blank in Table~\ref{tab:quantitative_comparison}.

  \paragraph{General-purpose language models.}
  These models are not trained for this task, so we provide the specification of our DSL and let them
  follow the same step-wise protocol as CAD-WAM.
  Each test model is produced in one conversation.
  The first turn contains the grammar of the action language, including the plane conventions of the
  source data, together with the description of the target model, and the model replies with one action
  per turn.
  After each reply the kernel executes the actions produced so far.
  If an action fails to execute, generation stops for that model, as in our own inference loop.
  Otherwise the resulting solid is rendered as three orthographic views, from the front, the top and the
  right, and returned with its measured bounding box; this rendering takes the place of the latent
  geometric state that CAD-WAM observes.
  A conversation ends when the model emits the end action, when an action fails, or after at most $12$
  actions, and the solid built so far is exported.
  A model is counted as invalid when no solid can be exported.
  Decoding uses temperature zero, and the views are rendered at $512$ pixels.
  \section{Evaluation Protocol}
  \label{app:eval}

  Published results are not comparable across papers, because each defines its metrics on a different
  normalization and a different number of samples.
  We therefore evaluate every system with a unified protocol, described below.
  Each metric follows the definition of the paper that introduced it, so that our values remain on the
  established scale.

  \paragraph{Meshing and sampling.}
  Let $P$ denote a predicted solid and $R$ the corresponding reference.
  Both are fitted into $[-1,1]^3$ by centering their bounding box and scaling by $2/\ell$, where $\ell$ is
  the longest side of that box.
  This fitting makes the meshing tolerances and the Boolean volumes comparable across models of different
  physical size.
  The solids are then tessellated with a linear deflection of $0.001$ and an angular deflection of $0.5$.
  From each mesh we draw $N=8{,}192$ samples uniformly with respect to area, which gives the point sets
  $X=\{\mathbf{x}_i\}$ and $Y=\{\mathbf{y}_j\}$ together with the normals of the faces they were sampled
  from.
  These sets are used for the Chamfer distance, the Hausdorff distance and the normal consistency.

  \paragraph{Normalization of the point sets.}
  Each metric applies the normalization of its original definition, which we keep so that our values can
  be read on the published scale.
  For the Chamfer and Hausdorff distances, each cloud is centered on its mean and divided by the range of
  its coordinates, taken over all three axes jointly, which places it in approximately $[-0.5,0.5]^3$.
  For the edge Chamfer distance, each cloud is centered on its mean and divided by the largest absolute
  coordinate, which places it in approximately $[-1,1]^3$.
  The two conventions differ by a factor of about two in linear distance, hence by about four in squared
  distance, but each is applied identically to every system, so the comparison between systems is
  unaffected.

  \paragraph{Chamfer distance: overall surface agreement.}
  The Chamfer distance measures how closely the two surfaces agree on average.
  With $d(\mathbf{x},Y)=\min_{\mathbf{y}\in Y}\lVert \mathbf{x}-\mathbf{y}\rVert$,
  \begin{equation}
  \mathrm{CD}(X,Y)=10^{3}\left(\frac{1}{|X|}\sum_{\mathbf{x}\in X} d(\mathbf{x},Y)^{2}
  +\frac{1}{|Y|}\sum_{\mathbf{y}\in Y} d(\mathbf{y},X)^{2}\right).
  \label{eq:cd}
  \end{equation}
  Both directions are summed and the result is scaled by $10^{3}$, following the convention of the
  compared work.
  Squaring the distances makes the measure sensitive to large local deviations, which is why we report the
  median as well as the mean.

  \paragraph{Hausdorff distance: worst-case deviation.}
    The Hausdorff distance reports the largest deviation between the two surfaces rather than the average,
    \begin{equation}
    \mathrm{HD}(X,Y)=10^{2}\max\Big\{\max_{\mathbf{x}\in X} d(\mathbf{x},Y),\;
    \max_{\mathbf{y}\in Y} d(\mathbf{y},X)\Big\}.
    \label{eq:hd}
    \end{equation}
    It is evaluated on the same point sets as the Chamfer distance and is not squared; the factor $10^{2}$
    is applied to every evaluated system.

  \paragraph{Normal consistency: orientation of the surfaces.}
  Normal consistency measures whether the surfaces agree in orientation and not only in position, which
  distinguishes a correctly reconstructed face from a nearby but differently oriented one.
  It is computed on the same $8{,}192$-point sets.
  Let $\mathbf{n}_{\mathbf{x}}$ be the normal of the face from which $\mathbf{x}$ was sampled and let
  $\pi_Y(\mathbf{x})$ denote the nearest neighbor of $\mathbf{x}$ in $Y$.
  Then
  \begin{equation}
  \mathrm{NC}(X,Y)=\frac{1}{2}\left(\frac{1}{|X|}\sum_{\mathbf{x}\in X}
  \big\langle \mathbf{n}_{\mathbf{x}}, \mathbf{n}_{\pi_Y(\mathbf{x})}\big\rangle
  +\frac{1}{|Y|}\sum_{\mathbf{y}\in Y}
  \big\langle \mathbf{n}_{\mathbf{y}}, \mathbf{n}_{\pi_X(\mathbf{y})}\big\rangle\right).
  \label{eq:nc}
  \end{equation}

  \paragraph{Edge Chamfer distance: fidelity of sharp features.}
  Mechanical parts are characterized by their edges, which a surface metric weights only weakly.
  This metric therefore compares the sharp edges alone, under its own sampling.
  Both shapes are sampled with $10^{5}$ points and normalized as described above.
  A point is assigned to an edge when a neighbor within a radius of $0.004$ has a normal satisfying
  $|\langle \mathbf{n},\mathbf{n}'\rangle|<0.2$.
  The edge Chamfer distance is Eq.~(\ref{eq:cd}) evaluated on the two sets of edge points, and we report
  its median.

  \paragraph{Intersection over union: agreement of the occupied volume.}
  This metric measures how much of the material of the two solids coincides, which a surface metric does
  not capture.
  Both solids are fitted into $[-1,1]^3$, and their intersection and union are obtained with Boolean
  operations of the kernel, evaluated with a fuzzy tolerance, while the volumes
  $V_P$, $V_R$, $V_{\cap}$ and $V_{\cup}$ are the exact volumes of the resulting solids,
  \begin{equation}
  \mathrm{IoU}(P,R)=\frac{V_{\cap}}{V_{\cup}}.
  \label{eq:iou}
  \end{equation}
  Boolean operations on nearly coincident faces are numerically delicate, so each result must satisfy the
  inclusion-exclusion identity
  \begin{equation}
  \big|\,V_{\cup}-(V_P+V_R-V_{\cap})\,\big|\le 10^{-3}\max\{V_{\cup},V_P,V_R\}.
  \label{eq:iou_check}
  \end{equation}
  We evaluate the Booleans with a fuzzy tolerance of $10^{-5}$, then without a tolerance, then with
  $10^{-4}$, and keep the first result that satisfies Eq.~(\ref{eq:iou_check}).
  Pairs for which no setting is consistent are excluded and their number is reported.

  \paragraph{Euler characteristic match: topological agreement.}
  This criterion checks whether the generated model has the same topology as the reference, for instance
  the same number of holes.
  It is evaluated only when both meshes are watertight, and equals one when their Euler characteristics
  agree after unreferenced vertices have been removed.

  \paragraph{Invalidity ratio: whether a usable model is produced at all.}
  A prediction is counted as invalid when its solid cannot be loaded, has zero volume, or cannot be
  meshed.
  The invalidity ratio is the share of such predictions among the models a system attempted,
  \begin{equation}
  \mathrm{IR}=1-\frac{|\{\text{valid predictions}\}|}{|\{\text{attempted models}\}|}.
  \label{eq:ir}
  \end{equation}
  A model counts as attempted when the system was run on it and, for the systems whose outputs we obtained
  from their authors, when an output was released for it.
  CAD-WAM and the zero-shot models attempt all 7{,}962 evaluation models, CAD-Coder attempts 7{,}958,
  cadrille 7{,}949, Text2CAD 7{,}841 and CADFS 7{,}212, as listed in Appendix~\ref{app:baselines}.
  All other metrics are computed on the valid predictions whose reference can be scored.
  They are therefore conditional on a system producing a solid at all, and must be read together with the
  invalidity ratio, which is defined over the same set of attempted models.

  \section{Textual State Variant}
  \label{app:text_state}

  This section specifies the variant of Section~\ref{sec:latent_vs_text}, in which the geometric state is
  expressed in text instead of being encoded into a latent vector.

  \paragraph{Architecture.}
  The variant uses the backbone without modification.
  Since the state is part of the text, neither the Geometric State Embedding nor the State Prediction Head
  is required, and the model has no additional module and no continuous input.
  It reads a prompt and produces text, exactly as a language model trained for any other generation task.
  The target combines the next action and the next state in a single JSON object, so the variant still
  predicts the state that its action produces, and it is trained with a cross-entropy loss over both.
  The weight $\lambda$ therefore does not apply.

\begin{figure}[ht]
  \centering
  \begin{tcolorbox}[sidebyside, righthand width=0.30\linewidth, sidebyside gap=6pt,
    colback=white, colframe=black!55, boxrule=0.5pt, arc=2pt,
    left=5pt, right=5pt, top=4pt, bottom=4pt]
\begin{lstlisting}[basicstyle=\ttfamily\scriptsize, columns=flexible]
{"solids": 1,
 "bbox": {"min": [-0.035139, 0.0, -0.023724],
          "max": [0.060353, 0.095213, 0.025922]},
 "volume": 0.00033102, "area": 0.046592,
 "faces": 7, "edges": 15, "vertices": 10,
 "face_list": [
   {"type": "cylinder",
    "center": [-0.0090427, 0.047607, 0.0],
    "axis": [0, 1, 0], "radius": 0.02006,
    "convex": false, "area": 0.012001},
   {"type": "plane",
    "center": [0.012607, 0.047607, -0.023724],
    "normal": [0, 0, -1], "area": 0.0090921},
   {"type": "plane",
    "center": [0.012607, 0.047607, 0.025922],
    "normal": [0, 0, 1], "area": 0.0090921},
   ...
 ]}
\end{lstlisting}
  \tcblower
  \centering\includegraphics[width=\linewidth]{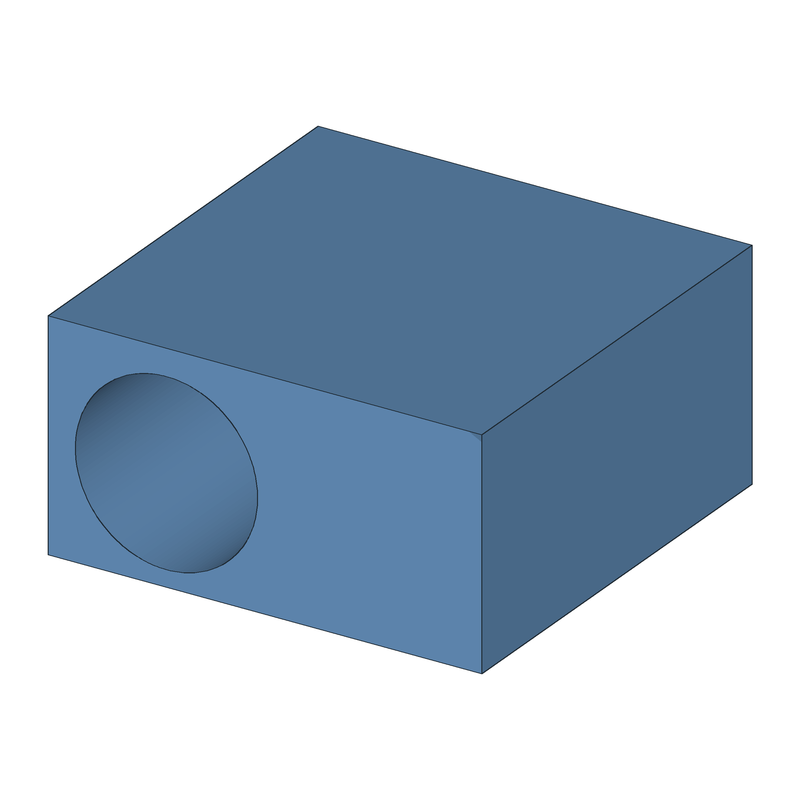}
  \end{tcolorbox}
  \caption{Textual state of an intermediate solid and the solid it describes. The part is a block with a
  cylindrical through-hole, which the state reports as six planar faces and one concave cylindrical face;
  three of the seven faces are shown.}
  \label{fig:text_state_example}
  \end{figure}

  \paragraph{Content of the textual state.}
  The textual state is a deterministic summary of the current solid, expressed in the coordinate frame and
  the units of the actions.
  It reports the number of solids, the bounding box, the volume, the surface area, and the numbers of
  faces, edges and vertices.
  It further lists the faces in decreasing order of area.
  Each entry gives the type of the face and the parameters of that type: a plane is described by its
  center and its normal, and a curved face by its center, its radius and its convexity, together with the
  axis of a cylinder, a cone or a torus, the half-angle of a cone, and the minor radius of a torus.
  Numbers are written with five significant digits, and the faces are ordered after rounding, so that a
  given geometry always yields the same text.
  The empty model is represented by \texttt{\{"solids": 0\}}.
  Figure~\ref{fig:text_state_example} illustrates the state of a solid with seven faces.

  \paragraph{Input and target.}
  The prompt contains the same instruction, the same description and the same progress information as in
  the main model, and the textual state replaces the geometric tokens.
  The target is a JSON object holding the next action and the state that this action produces.
  At the end of a construction, the action is the end action and the state is left unchanged.

  \paragraph{Training and cost.}
  The variant is trained on the same records, with the same backbone and the same optimization settings as
  the main model.
  The textual state dominates the input and the target: it accounts for about four fifths of the tokens of
  a target, and the length of a record grows with the number of faces of the current solid.

  \section{Effect of the State Prediction Weight}
  \label{app:lambda}
 \begin{table}[t]
  \centering
  \caption{
  Ablation study on the state prediction loss weight $\lambda$.
  CD, HD, and NC are reported as mean values.
  }
  \label{tab:state_prediction_loss_weight}

  \small
  \setlength{\tabcolsep}{3.2pt}
  \renewcommand{\arraystretch}{1.15}
  \renewcommand{\tabularxcolumn}[1]{m{#1}}

  \begin{tabularx}{\linewidth}{
      @{}
      >{\centering\arraybackslash}m{0.25\linewidth}
      *{5}{>{\centering\arraybackslash}X}
      @{}
  }
      \toprule

      \textbf{Metric}
      & $\lambda=0$
      & $\lambda=0.01$
      & $\lambda=0.03$
      & $\lambda=0.05$
      & $\lambda=0.1$
      \\

      \midrule

      IR (\%) $\downarrow$
      & 0.779
      & 0.691
      & \textbf{0.666}
      & 0.791
      & 0.804
      \\

      CD $\downarrow$
      & 1.316
      & 1.229
      & 1.142
      & 1.232
      & \textbf{1.107}
      \\

      HD $\downarrow$
      & 4.264
      & 4.203
      & \textbf{4.134}
      & 4.163
      & 4.184
      \\

      NC $\uparrow$
      & 0.902
      & 0.902
      & \textbf{0.903}
      & 0.903
      & 0.903
      \\

      ECD $\downarrow$
      & 7.374
      & 7.251
      & 7.313
      & \textbf{7.189}
      & 7.196
      \\

      IoU $\uparrow$
      & 0.960
      & 0.961
      & \textbf{0.964}
      & 0.963
      & 0.962
      \\

      Euler match (\%) $\uparrow$
      & 95.404
      & 95.333
      & \textbf{95.850}
      & 95.797
      & 95.629
      \\

      \bottomrule
  \end{tabularx}

  \end{table}

  The weight $\lambda$ of Eq.~(\ref{eq:loss}) sets the relative contribution of the auxiliary state
  objective.
  We train one model per value and evaluate all of them with the protocol of Appendix~\ref{app:eval};
  every other setting is identical.

  As reported in Table~\ref{tab:state_prediction_loss_weight}, every positive weight improves upon
  $\lambda=0$ across the metrics that characterize geometric quality, namely the Chamfer and Hausdorff
  distances, the normal consistency and the edge Chamfer distance, and the intersection over union
  improves as well.
  The remaining metrics vary within a narrow band around their values at $\lambda=0$.
  The differences among the positive weights are small, so state prediction acts as an auxiliary signal
  whose benefit does not depend on a precise weight.
  We adopt $\lambda=0.03$ in all other experiments.
\begin{figure}[t]
    \centering
    \includegraphics[width=\linewidth]{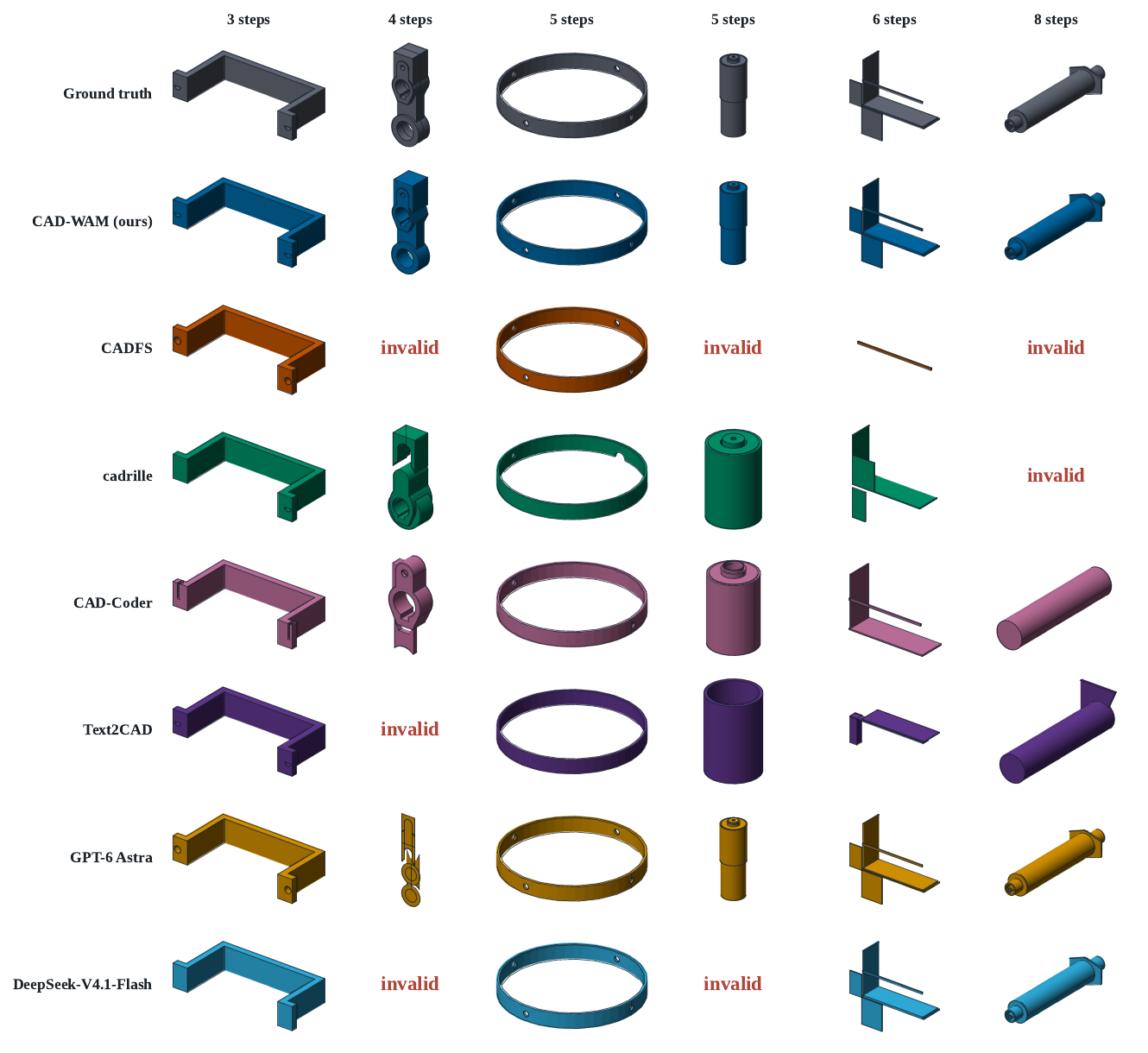}
    \caption{Predictions of each system on six test models, ordered by the number of
    construction steps of the reference. \emph{invalid} marks a prediction that does
    not build a valid solid.}
    \label{fig:gallery}
  \end{figure}

  \section{Qualitative Examples}\label{app:qualitative}

  Figure~\ref{fig:gallery} compares the solids produced by every system on six models
  of the evaluation set whose references require three to eight construction steps.
  Each row corresponds to one system and each column to one model, with the reference
  in the first row. Every solid is fitted to $[-1,1]^3$, the normalization under which
  the metrics of Appendix~\ref{app:eval} are computed, and all cells of a column are
  rendered from a single camera, so that the predictions of different systems can be
  compared directly.

  On the model with the fewest steps every system recovers the intended shape, which
  is the regime that the median over the full evaluation set reflects. As the number
  of steps increases, the predictions of the baselines depart from the reference in
  two ways: they collapse to a single primitive that matches the overall proportions
  but none of the detail, such as a plain cylinder in place of a stepped shaft or a
  flat plate in place of a finned bracket, or they fail to produce a valid solid at
  all. CAD-WAM builds a valid solid for all six models and reproduces the features
  introduced by the later steps, with a Chamfer distance below $0.15$ throughout.
  These examples illustrate at the level of individual models the behavior observed
  in aggregate on the full evaluation set: the advantage of predicting the geometric
  state grows with the number of operations a model requires.
\section{Limitations}
  \label{app:limitations}
  Our evaluation is limited to the sketch-and-extrude programs in DeepCAD. Extending CAD-WAM to richer operations and longer    
  construction sequences would help assess its applicability to more demanding modeling tasks. The state prediction head currently supplies supervision only. Since it estimates the outcome of an
  action before execution, it can equally serve as a look-ahead signal: candidate actions are scored
  against their predicted states, and the model commits to one only after that comparison.

\end{document}